\documentclass[journal=jcisd8,manuscript=article]{achemso}

\usepackage{graphicx}
\usepackage{booktabs}
\usepackage{amsmath}
\usepackage{amssymb}
\usepackage{mathtools}
\usepackage{bm}
\usepackage{float}
\usepackage{xurl}

\author{Carles Navarro}
\email{c.navarro@acellera.com}
\affiliation[Acellera Labs]{Acellera Labs, C/ Doctor Trueta 183, 08005 Barcelona, Spain}
\alsoaffiliation[Universitat Pompeu Fabra]{Computational Science Laboratory, Universitat Pompeu Fabra, PRBB, C/ Doctor Aiguader 88, 08003 Barcelona, Spain}

\author{Philipp Tholke}
\affiliation[Acellera Labs]{Acellera Labs, C/ Doctor Trueta 183, 08005 Barcelona, Spain}

\author{Gianni de Fabritiis}
\email{g.defabritiis@acellera.com}
\affiliation[Acellera Labs]{Acellera Labs, C/ Doctor Trueta 183, 08005 Barcelona, Spain}
\alsoaffiliation[Universitat Pompeu Fabra]{Computational Science Laboratory, Universitat Pompeu Fabra, PRBB, C/ Doctor Aiguader 88, 08003 Barcelona, Spain}
\alsoaffiliation[ICREA]{Instituci\'{o} Catalana de Recerca i Estudis Avan\c{c}ats (ICREA), Passeig Llu\'{i}s Companys 23, 08010 Barcelona, Spain}

\title{Structure-guided molecular design with contrastive 3D protein–ligand learning}


\begin{document}

%%%%%%%%%%%%%%%%%%%%%%%%%%%%%%%%%%%%%%%%%%%%%%%%%%%%%%%%%%%%%%%%%%%%%
%% Table of Contents Entry
%%%%%%%%%%%%%%%%%%%%%%%%%%%%%%%%%%%%%%%%%%%%%%%%%%%%%%%%%%%%%%%%%%%%%
%\begin{tocentry}
%\includegraphics[width=8.25cm,height=4.45cm,keepaspectratio]{imgs/figure_1.png}
%\end{tocentry}

%%%%%%%%%%%%%%%%%%%%%%%%%%%%%%%%%%%%%%%%%%%%%%%%%%%%%%%%%%%%%%%%%%%%%
%% Abstract
%%%%%%%%%%%%%%%%%%%%%%%%%%%%%%%%%%%%%%%%%%%%%%%%%%%%%%%%%%%%%%%%%%%%%
\begin{abstract}
Structure-based drug discovery faces the dual challenge of accurately capturing 3D protein-ligand interactions while navigating ultra-large chemical spaces to identify synthetically accessible candidates. In this work, we present a unified framework that addresses these challenges by combining contrastive 3D structure encoding with autoregressive molecular generation conditioned on commercial compound spaces. First, we introduce an SE(3)-equivariant transformer that encodes ligand and pocket structures into a shared embedding space via contrastive learning, achieving competitive results in zero-shot virtual screening. Second, we integrate these embeddings into a multimodal Chemical Language Model (MCLM). The model generates target-specific molecules conditioned on either pocket or ligand structures, with a learned dataset token that steers the output toward targeted chemical spaces, yielding candidates with favorable predicted binding properties across diverse targets.
\end{abstract}

%%%%%%%%%%%%%%%%%%%%%%%%%%%%%%%%%%%%%%%%%%%%%%%%%%%%%%%%%%%%%%%%%%%%%
%% Introduction
%%%%%%%%%%%%%%%%%%%%%%%%%%%%%%%%%%%%%%%%%%%%%%%%%%%%%%%%%%%%%%%%%%%%%
\section{Introduction}
\label{sec:intro}

Virtual screening, a critical step in drug discovery, is often described as a "needle in a haystack" problem: identifying biologically active small molecules against a specific protein target from astronomically large chemical spaces \cite{Raghuraman2006needle,polishchuk2013estimation,bohacek1996art}. While advances in computational power and algorithms have made in-silico virtual screening increasingly effective \cite{shoichet2004virtual,Lyu2019}, the sheer scale of modern chemical libraries continues to outpace available methods. The core objective is to identify molecules from an explicit or implicit chemical space that are likely to bind to a protein target and form a stable complex. These chemical spaces range from curated catalogs of physical stock to massive, enumerated libraries of synthetically accessible compounds, such as the Enamine REAL Space, which currently exceeds tens of billions of molecules and continues to grow rapidly \cite{grygorenko2020generating,warr2022exploration}. As these libraries expand, exhaustive screening using traditional structure-based techniques, such as molecular docking \cite{sivula2023machine}, or even faster ligand-based approaches \cite{hawkins2007comparison}, is becoming increasingly impractical and computationally intractable.

In this work, we present a unified framework that overcomes these scale and applicability limitations by combining contrastive 3D protein-ligand encodings with a multimodal Chemical Language Model (MCLM). First, we introduce an $SE(3)$-equivariant transformer encoder (SET) that maps ligand and pocket 3D structures into a shared embedding space via a contrastive objective. This produces compact representations that effectively capture the geometric and chemical features relevant to binding compatibility, the quality of which we demonstrate through zero-shot virtual screening on the challenging LIT-PCBA benchmark. Second, we integrate these structural embeddings into an autoregressive CLM. Crucially, generation is conditioned on both the 3D embedding and a learned dataset token that steers the output toward a specific chemical space. This integrated design enables the de novo generation of target-specific molecules that closely resemble commercial libraries, retaining the high efficiency of generative approaches while directly leveraging 3D structural information from protein targets to ensure synthetic accessibility.

Our approach builds upon and addresses the limitations of several recent deep learning paradigms for in silico virtual screening and molecular generation. Contrastive learning methods have recently enabled the fast vector retrieval of binding-compatible molecules by encoding proteins and ligands into shared embedding spaces, bypassing the need to exhaustively dock ultra-large libraries \cite{drugclip}. Separately, Chemical Language Models (CLMs) represent molecules as SMILES token sequences and generate novel structures through autoregressive learning \cite{segler2018}, offering an efficient means of exploring implicit chemical spaces far larger than any enumerated library \cite{arus2019exploring, du2024machine}. However, standard CLMs do not incorporate 3D structural information from protein targets and frequently generate molecules that are not commercially available or synthetically accessible, often requiring additional optimization steps such as reinforcement learning or docking-based scoring to produce target-specific candidates \cite{reinvent}. More recently, multimodal architectures such as COATI \cite{coati} have begun to bridge this gap by combining a contrastive encoder with an autoregressive decoder to generate molecules conditioned on embeddings. Generative models that directly operate on 3D molecular structures have also emerged as an alternative. In particular, diffusion-based approaches that generate molecules as atom point clouds within protein pocket cavities have gained considerable attention \cite{pocket2mol, targetdiff, diffsbdd, pilot}. Nevertheless, these 3D models face notable limitations: their training is restricted by the scarcity of high-quality 3D ligand-protein structural data, generated geometries often require post-processing to correct valency errors or strained conformations, and because they rely strictly on experimentally determined structures, they cannot readily learn or be steered toward the distribution of a desired commercial chemical space.

%%%%%%%%%%%%%%%%%%%%%%%%%%%%%%%%%%%%%%%%%%%%%%%%%%%%%%%%%%%%%%%%%%%%%
%% Methods
%%%%%%%%%%%%%%%%%%%%%%%%%%%%%%%%%%%%%%%%%%%%%%%%%%%%%%%%%%%%%%%%%%%%%
\section{Methods}
\label{sec:methods}
\subsection{Scalable Equivariant Transformer Encoder}
\label{sec:scet}

To efficiently process 3D atomic point clouds, we introduce the Scalable Equivariant Transformer (SET), an architecture built upon the standard Transformer that incorporates SE(3)-equivariance. SET extends distance-aware dot-product attention\cite{brehmer2023geometric} to also account for reflections, enabling the processing of both scalar and vector features within a unified self-attention framework that leverages highly optimized dot-product attention implementations. This design preserves the scalability of standard Transformers while ensuring representational consistency under physical transformations, enabling the training of large models on extensive datasets. For a detailed description of the SET architecture, see Section~S1 in the Supporting Information.

The model takes as input the atomic numbers $\bm{z}_i \in \mathbb{R}$ and the 3D coordinates $\bm{p}_i \in \mathbb{R}^3$ of all $N$ heavy atoms in the system. We prepend a learnable vector $\bm{e}_{\text{cls}} \in \mathbb{R}^d$ as a virtual $0$-th atom at the center of mass, $\bm{p}_0 = \tfrac{1}{N}\sum_{i=1}^{N} \bm{p}_i$. This token serves as a global aggregator, pooling information across all atoms:
\begin{equation}
\bm{h}^{(c)} = \bigl[\mathrm{SET}\bigl(\bm{e}_{\text{cls}}, \{\bm{z}_i, \bm{p}_i\}_{i=1}^{N}\bigr)\bigr]_0 \in \mathbb{R}^d, \qquad c \in \{\text{l}, \text{p}\},
\label{eq:class_embedding}
\end{equation}
with $c \in \{\text{l}, \text{p}\}$ representing ligand and pocket, respectively. A one-layer MLP $g(\cdot)$ then projects $\bm{h}^{(c)}$ into the final multimodal embedding space, yielding $\bm{x}^{(c)} = g(\bm{h}^{(c)}) \in \mathbb{R}^{d'}$.

\subsection{Contrastive Learning}
\label{sec:contrastive}
\begin{figure}[H]
    \centering
    \includegraphics[width=\linewidth]{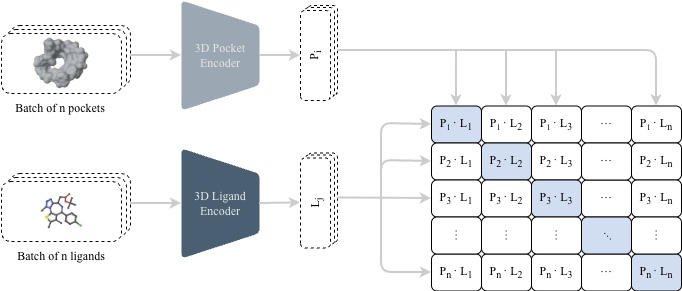}
    \caption{Contrastive training of the shared ligand-pocket embedding space. A batch of $n$ pairs is encoded by the ligand and pocket branches into projections $L_i$ and $P_i$; the CF-InfoNCE loss aligns positive pairs along the diagonal of the pairwise similarity matrix.}
    \label{fig:clipp_arch}
\end{figure}

Two SET encoders are jointly trained with a contrastive objective that aligns ligand and pocket representations of the same complex, following the CLIP\cite{clip} framework of treating off-diagonal pairs in the similarity matrix as negatives (Figure~\ref{fig:clipp_arch}). We refer to the resulting pair of encoders as \textbf{CLIPP-SET} (Contrastive Ligand-Pocket Pretraining with SET). In drug discovery, however, multiple ligands can bind to the same pocket, and while datasets contain millions of unique ligand-pocket pairs, the number of unique pockets is comparatively small. A batch therefore often contains repeated pockets with more than one active ligand, and not all off-diagonal pairs can be treated as true negatives. To address this, we adopt the Collision-Free InfoNCE (CF-InfoNCE) loss,\cite{wang2024uniclip} a variant of InfoNCE\cite{oord2018representation} that handles pocket collisions by dynamically selecting the ligand with the strongest binding affinity among those sharing the same pocket as the true positive for the pocket-to-ligand direction.

Specifically, for a mini-batch of $N$ ligand-pocket pairs $\{(\bm{x}^l_i, \bm{x}^p_i)\}_{i=1}^N$, the pocket-to-ligand loss for pocket $\bm{x}^p_i$ is defined as:
\begin{equation}
\mathcal{L}^p_i =
- \log \frac{\exp(s(\bm{x}^p_i, \bm{x}^l_{j^*}) / \tau)}
{\sum_{j=1}^N \exp(s(\bm{x}^p_i, \bm{x}^l_j) / \tau)},
\label{eq:pocket_loss}
\end{equation}
where $j^* = \arg\max_{j \in C_i} v_{j,i}$, $C_i = \{j : \text{pocket}_j = \text{pocket}_i\}$ denotes the set of indices sharing the same pocket as sample $i$, and $v_{j,i}$ is the binding affinity of the ligand-pocket pair $(\bm{x}^l_j, \bm{x}^p_i)$. When $|C_i| = 1$ (no collision), $j^* = i$ and this reduces to the standard InfoNCE loss.

For the ligand-to-pocket direction, the diagonal pairs remain the true positives, since ligand collisions are rare in practice:
\begin{equation}
\mathcal{L}^l_i =
- \log \frac{\exp(s(\bm{x}^l_i, \bm{x}^p_i) / \tau)}
{\sum_{j=1}^N \exp(s(\bm{x}^l_i, \bm{x}^p_j) / \tau)}.
\label{eq:ligand_loss}
\end{equation}

The total loss for the mini-batch is:
\begin{equation}
\mathcal{L} = \frac{1}{2} \sum_{i=1}^N \big(\mathcal{L}^p_i + \mathcal{L}^l_i\big).
\label{eq:total_loss}
\end{equation}

\subsection{Multimodal Chemical Language Model}
\label{sec:mm_clm}

\begin{figure}[H]
    \centering
    \includegraphics[width=0.85\linewidth]{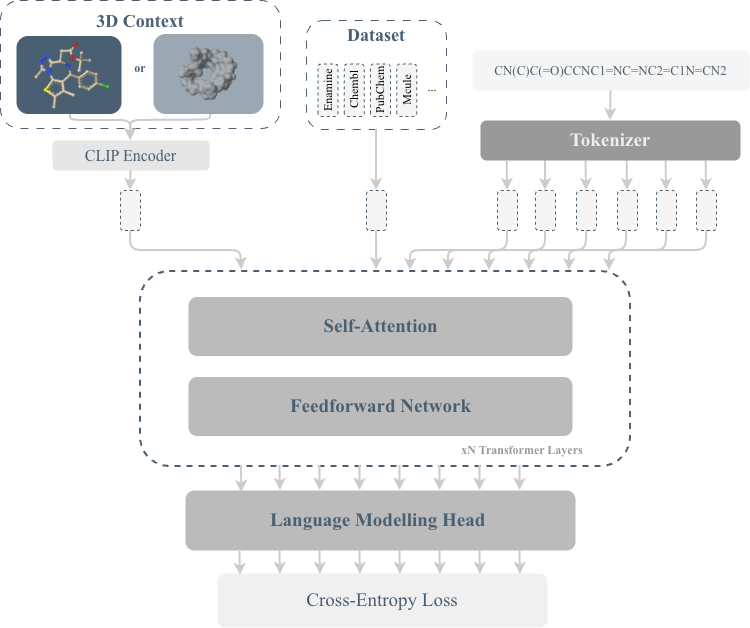}
    \caption{MCLM architecture. A ligand or pocket 3D context is encoded by the frozen contrastive encoder; the projection is concatenated with a dataset token and the tokenized SMILES prefix, and the sequence is processed by an autoregressive decoder trained with cross-entropy on the SMILES token.}
    \label{fig:mclm_arch}
\end{figure}

We integrate the contrastive 3D structure encoder with a generative Chemical Language Model\cite{segler2018} (CLM) to form a multimodal framework for target-specific molecular generation (Figure~\ref{fig:mclm_arch}). The decoder is an autoregressive model built on the Llama2 architecture,\cite{touvron2023llama} where the learned structural embedding is projected and prepended to the input tokens, conditioning generation on 3D structural features in a manner analogous to multimodal models that prepend visual tokens to language model inputs.\cite{alayrac2022flamingo,li2023blip} In addition to the contrastive embedding vector $\bm{x}$ produced by the SET encoder for a given ligand or pocket, the input sequence includes a  dataset token $[\bm{ds}]$ indicating the chemical space from which the training sample originates. At inference time, selecting a specific dataset token steers generation toward that chemical domain.

The final input sequence $\bm{H}$ to the language model is:
\begin{equation}
\bm{H} = \bigl[\,[\bm{ds}],\,\bm{x}\mathbf{W}_P,\, \bm{s}_1,\, \bm{s}_2,\, \ldots,\, \bm{s}_n\bigr],
\label{eq:input_seq}
\end{equation}
where $\bm{s}_i$ denotes the standard token embeddings of the partial SMILES sequence and $\mathbf{W}_P$ is a trainable linear projection. The model is trained with the standard autoregressive loss,\cite{bengio2000neural} minimizing the negative log-likelihood of the correct SMILES token at each timestep:
\begin{equation}
\mathcal{L}_\text{NLL} = -\sum_{t=1}^m \log p_\theta\bigl(\bm{s}_t \,\big|\, \bm{s}_{<t}, [\bm{ds}], \bm{x}\mathbf{W}_P\bigr).
\label{eq:nll_loss}
\end{equation}

\subsection{Datasets}
\label{sec:datasets}

\paragraph{Conformer Dataset.}
We assembled a conformer dataset of approximately 287 million conformer-SMILES pairs from five public and commercial databases: PubChem\cite{pubchem} (97.7M conformers), the Enamine Diversity Set (96.8M), Mcule (86.5M), ChEMBL\cite{chembl} (4.3M), and GEOM-drugs\cite{geom_drugs} (1.9M). PubChem and GEOM-drugs provided pre-existing 3D structures, while conformers for ChEMBL, Mcule, and the Enamine Diversity Set were generated using the ETKDG method in RDKit and minimized with the MMFF force field. Each entry consists of the conformer's atomic numbers, coordinates, canonical SMILES, and a source label indicating the database of origin.

\paragraph{SIU Dataset.}
The contrastive encoder was trained on the SIU dataset,\cite{siu} a collection of 1,291,362 molecule-pocket pairs spanning 9,662 unique pockets and 214,686 unique molecules. Binding poses are generated via consensus docking, and each pair is annotated with a measured binding affinity (IC$_{50}$, K$_i$, EC$_{50}$, or K$_d$).

\paragraph{ProFSA Dataset.}
The pocket encoder was pretrained on the ProFSA dataset,\cite{profsa} which provides approximately 5.5 million fragment-pocket pairs derived from 53,824 unique PDB structures clustered at 70\% sequence identity. 

\paragraph{Evaluation Datasets.}
We used two datasets for evaluation. The LIT-PCBA benchmark\cite{litpcba} served as both the virtual screening and generative evaluation dataset, comprising 15 targets with 7,761 active compounds and 382,674 unique inactives. The Enamine REAL database was used both as a search space for nearest-neighbor retrieval experiments and as the target chemical domain for conditioned generation.

\subsection{Training}
\label{sec:training}

\paragraph{Encoder Pretraining.}
Both the ligand and pocket encoders share the same SET architecture with a 256-dimensional projection head. Each encoder is pretrained independently using a masked language modeling (MLM) objective\cite{devlin2019bert} over its respective domain: the ligand encoder on the conformer dataset described below and the pocket encoder on the ProFSA dataset.\cite{profsa} During pretraining, 20\% of input atoms are selected for prediction, of which 80\% are masked, 10\% are replaced with a random atom type, and 10\% are left unchanged. Full hyperparameter details are provided in the Supporting Information.

\paragraph{Contrastive Training.}
The pretrained ligand and pocket encoders are jointly trained on the SIU dataset\cite{siu} using the CF-InfoNCE loss defined in Eqs.~\ref{eq:pocket_loss}--\ref{eq:total_loss} with a learnable temperature parameter $\tau$.\cite{clip} When multiple binding poses are available for a given pair, one conformer is selected at random during training. Full hyperparameter details are provided in the Supporting Information.

\paragraph{Multimodal CLM Training.}
The MCLM is trained on the conformer dataset using a dataset token indicating the source chemical space and a pre-computed embedding from the frozen contrastive encoder, both prepended to the token sequence as defined in Eq.~\ref{eq:input_seq}. Each 256-dimensional embedding is projected to the CLM hidden dimension via a trainable linear layer. The model is trained with the autoregressive loss in Eq.~\ref{eq:nll_loss}. Full hyperparameter details are provided in the Supporting Information.

\subsection{Inference and Generation}
\label{sec:inference}
At inference time, molecules are generated conditioned on either a protein pocket or a ligand structure. The input structure is first encoded by the corresponding branch of the frozen contrastive encoder into a 256-dimensional embedding. For ligand inputs provided as SMILES strings, a 3D conformer is generated using RDKit and hydrogen atoms are removed prior to encoding. The resulting embedding, together with a dataset token specifying the target chemical space, is prepended to the decoder input as defined in Eq.~\ref{eq:input_seq}. Molecules are then generated autoregressively by sampling from the output distribution until an end-of-sequence token is produced or a maximum sequence length is reached.

%%%%%%%%%%%%%%%%%%%%%%%%%%%%%%%%%%%%%%%%%%%%%%%%%%%%%%%%%%%%%%%%%%%%%
%% Results
%%%%%%%%%%%%%%%%%%%%%%%%%%%%%%%%%%%%%%%%%%%%%%%%%%%%%%%%%%%%%%%%%%%%%
\section{Results}
\label{sec:results}

\subsection{Contrastive Embedding Performance on LIT-PCBA}
\label{sec:litpcba}

We evaluate the learned pocket embeddings on the LIT-PCBA virtual screening benchmark. This setting directly tests the quality of the pocket representations in a zero-shot regime, without any target-specific fine-tuning. The evaluation metrics include the area under the receiver operating characteristic curve (AUROC), which measures the overall ranking of actives against decoys; the Boltzmann-enhanced discrimination of ROC (BEDROC), a ROC variant that weights the top of the ranked list through an exponential decay to emphasize early recognition; and the enrichment factor (EF) at 0.5\%, 1\%, and 5\%, defined as the fold increase in actives recovered within the top $x\%$ of the ranked list relative to a random baseline.

Since our model was trained on binding conformations, predictions are sensitive to conformer variability when computing cosine similarity between pocket and ligand embeddings. To account for this, we generate 64 conformations per molecule using RDKit and select the one with the highest similarity as the final score. Table~\ref{tab:litpcba} reports results alongside several structure-based baselines that also screen using only pocket information. CLIPP-SET achieves the highest BEDROC, EF(0.5\%), and EF(1\%) across all baselines, but falls behind at the broader EF(5\%) threshold and on AUROC, where some docking-based methods perform better. 

\begin{table}[H]
    \centering
    \caption{Results on LIT-PCBA virtual screening benchmark using pocket-based screening. Baseline results are reproduced from Table~3 of Gao et al.\cite{drugclip} Best results in each column are shown in bold.}
    \label{tab:litpcba}
    \small
    \begin{tabular}{l@{\hskip 8pt}c@{\hskip 8pt}c@{\hskip 8pt}c@{\hskip 8pt}c@{\hskip 8pt}c}
        \toprule
        Method & AUROC (\%) & BEDROC (\%) & EF (0.5\%) & EF (1\%) & EF (5\%) \\
        \midrule
        Surflex\cite{surflex36}   & 51.47 & --   & --   & 2.50 & --   \\
        Glide-SP\cite{glide11}    & 53.15 & 4.00 & 3.17 & 3.41 & 2.01 \\
        Planet\cite{planet48}     & 57.31 & --   & 4.64 & 3.87 & \textbf{2.43} \\
        Gnina\cite{gnina27}       & \textbf{60.93} & 5.40 & --   & 4.63 & --   \\
        DeepDTA\cite{deepdta30}   & 56.27 & 2.53 & --   & 1.47 & --   \\
        BigBind\cite{bigbind2}    & 60.80 & --   & --   & 3.82 & --   \\
        DrugCLIP\cite{drugclip}   & 57.17 & 6.23 & 8.56 & 5.51 & 2.27 \\
        \midrule
        CLIPP-SET (pocket)        & 53.76 & \textbf{6.55} & \textbf{9.66} & \textbf{5.63} & 2.06 \\
        \bottomrule
    \end{tabular}
\end{table}

\subsection{Multimodal Search over Billion-Scale Chemical Libraries}
\label{sec:enamine}
To validate the learned pocket embeddings on a realistic drug discovery task, we conducted a large-scale search experiment on the Enamine REAL database (5.9 billion compounds). For each of the 15 LIT-PCBA targets, we retrieved the top 100 molecules using the CLIPP-SET pocket embedding and compared the results using Morgan fingerprint (FP) search, a 2D similarity baseline that requires a known reference ligand. LIT-PCBA provides multiple crystal structures per target; we selected a single representative PDB for each target to define the pocket query. Figure~\ref{fig:search_overview}c outlines both pipelines: CLIPP-SET encodes 3D coordinates with the SE(3)-equivariant transformer and retrieves neighbors by cosine similarity in the contrastive embedding space, whereas Morgan FP computes 2D circular fingerprints and retrieves neighbors by Tanimoto similarity. We predicted binding poses and affinities for each retrieved set using Boltz-2,\cite{boltz2} reporting affinity probability and predicted pIC$_{50}$, and measured internal chemical diversity across the retrieved molecules.

Table~\ref{tab:search_comparison_table} reports per-target results. Morgan FP search attains a higher mean affinity probability (0.502 vs.\ 0.451), while the pocket search achieves the best predicted pIC$_{50}$ on 13 of 15 targets and the highest diversity on 14 of 15 targets, with a mean pIC$_{50}$ of 5.71 compared to 5.45 for Morgan FP search. Crucially, this is achieved without any reference ligand, relying solely on the 3D geometry of the protein pocket, whereas Morgan FP retrieves close analogs of a known active, resulting in lower chemical diversity (0.587 vs.\ 0.788) and molecules that remain structurally close to existing compounds. These results suggest that the contrastive pocket representations capture binding-relevant geometric features of the target site, enabling the identification of novel and diverse candidates suitable for downstream drug discovery campaigns.

\begin{table}[H]
    \centering
    \caption{Per-target predicted binding affinity and chemical diversity for 100 molecules retrieved from the Enamine REAL database. CLIPP-SET: pocket-based search; Morgan: fingerprint similarity search using a known reference ligand. Best result per column for each target in bold.}
    \label{tab:search_comparison_table}
    \resizebox{\textwidth}{!}{%
    \begin{tabular}{l cc cc cc}
        \toprule
        & \multicolumn{2}{c}{Aff.\ Prob.\ $\uparrow$} & \multicolumn{2}{c}{Pred.\ pIC50 $\uparrow$} & \multicolumn{2}{c}{Diversity $\uparrow$} \\
        \cmidrule(lr){2-3} \cmidrule(lr){4-5} \cmidrule(lr){6-7}
        Target & CLIPP-SET & Morgan & CLIPP-SET & Morgan & CLIPP-SET & Morgan \\
        \midrule
        MTORC1 (1fap)   & 0.421 & \textbf{0.464} & \textbf{5.50} & 5.40 & \textbf{0.838} & 0.651 \\
        ESR1\_ago (1l2i) & 0.358 & \textbf{0.698} & 5.38 & \textbf{5.55} & \textbf{0.810} & 0.596 \\
        MAPK1 (1pme)    & \textbf{0.433} & 0.427 & \textbf{5.46} & 5.01 & \textbf{0.751} & 0.673 \\
        ESR1\_ant (1xp1) & 0.517 & \textbf{0.632} & \textbf{5.78} & 5.42 & \textbf{0.811} & 0.636 \\
        PPARG (1zgy)    & \textbf{0.459} & 0.217 & \textbf{5.65} & 4.60 & 0.639 & \textbf{0.667} \\
        GBA (2v3d)      & 0.496 & \textbf{0.624} & \textbf{5.67} & 5.56 & \textbf{0.862} & 0.685 \\
        TP53 (2vuk)     & 0.552 & \textbf{0.758} & \textbf{5.91} & 5.74 & \textbf{0.809} & 0.460 \\
        VDR (3a2i)      & \textbf{0.471} & 0.436 & \textbf{6.13} & 5.86 & \textbf{0.784} & 0.658 \\
        PKM2 (3gqy)     & 0.293 & \textbf{0.489} & \textbf{5.17} & 5.10 & \textbf{0.582} & 0.404 \\
        ADRB2 (3p0g)    & \textbf{0.703} & 0.617 & \textbf{6.54} & 6.22 & \textbf{0.859} & 0.511 \\
        IDH1 (4i3k)     & \textbf{0.409} & 0.381 & \textbf{5.05} & 4.88 & \textbf{0.833} & 0.589 \\
        ALDH1 (4wpn)    & 0.435 & \textbf{0.470} & \textbf{6.04} & 4.98 & \textbf{0.781} & 0.577 \\
        FEN1 (5fv7)     & 0.299 & \textbf{0.349} & \textbf{5.24} & 4.83 & \textbf{0.824} & 0.510 \\
        KAT2A (5h86)    & \textbf{0.284} & 0.113 & \textbf{5.83} & 5.81 & \textbf{0.794} & 0.621 \\
        OPRK1 (6b73)    & 0.629 & \textbf{0.856} & 6.32 & \textbf{6.81} & \textbf{0.841} & 0.561 \\
        \midrule
        Mean & 0.451 & \textbf{0.502} & \textbf{5.71} & 5.45 & \textbf{0.788} & 0.587 \\
        \bottomrule
    \end{tabular}%
    }
\end{table}

\subsection{Ligand-Based Search and Scaffold Hopping}
\label{sec:scaffold_hopping}

Because both ligands and pockets are encoded into the same embedding space, the framework naturally supports a second mode of search: given a known active ligand, one can retrieve molecules with similar 3D embeddings directly. This ligand-based search is particularly useful when a hit compound is already known and the goal is to identify structurally distinct alternatives that preserve binding-relevant 3D features.

\begin{figure}[H]
    \centering
    \includegraphics[width=\linewidth]{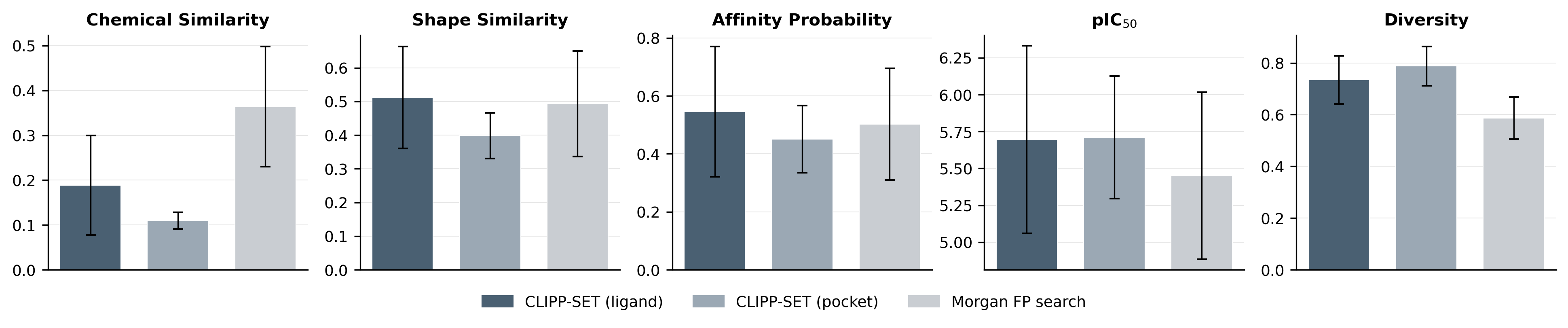}
    \caption{Mean search metrics across 15 LIT-PCBA targets for CLIPP-SET ligand search, CLIPP-SET pocket search, and Morgan FP search. Binding affinity predicted by Boltz-2.}
    \label{fig:search_barplot}
\end{figure}

We repeated the Enamine REAL search experiment using the CLIPP-SET ligand embedding of the co-crystal ligand of each of the 15 LIT-PCBA targets. In addition to the binding affinity and diversity metrics used in the pocket search, we evaluated the retrieved molecules with two ligand-based metrics: 3D shape and pharmacophoric similarity to the reference ligand,\cite{roshambo2} and 2D chemical similarity, measured as the Tanimoto coefficient between Morgan fingerprints of the retrieved and reference molecules. Figure~\ref{fig:search_barplot} compares the mean metrics across the three search strategies. The ligand search achieves the highest shape similarity and affinity probability while maintaining low chemical similarity to the reference ligand, indicating that it retrieves molecules that are 3D-compatible but chemically distinct from the query, unlike Morgan FP search, which retrieves close 2D analogs.

\begin{figure}[H]
    \centering
    \includegraphics[width=\linewidth]{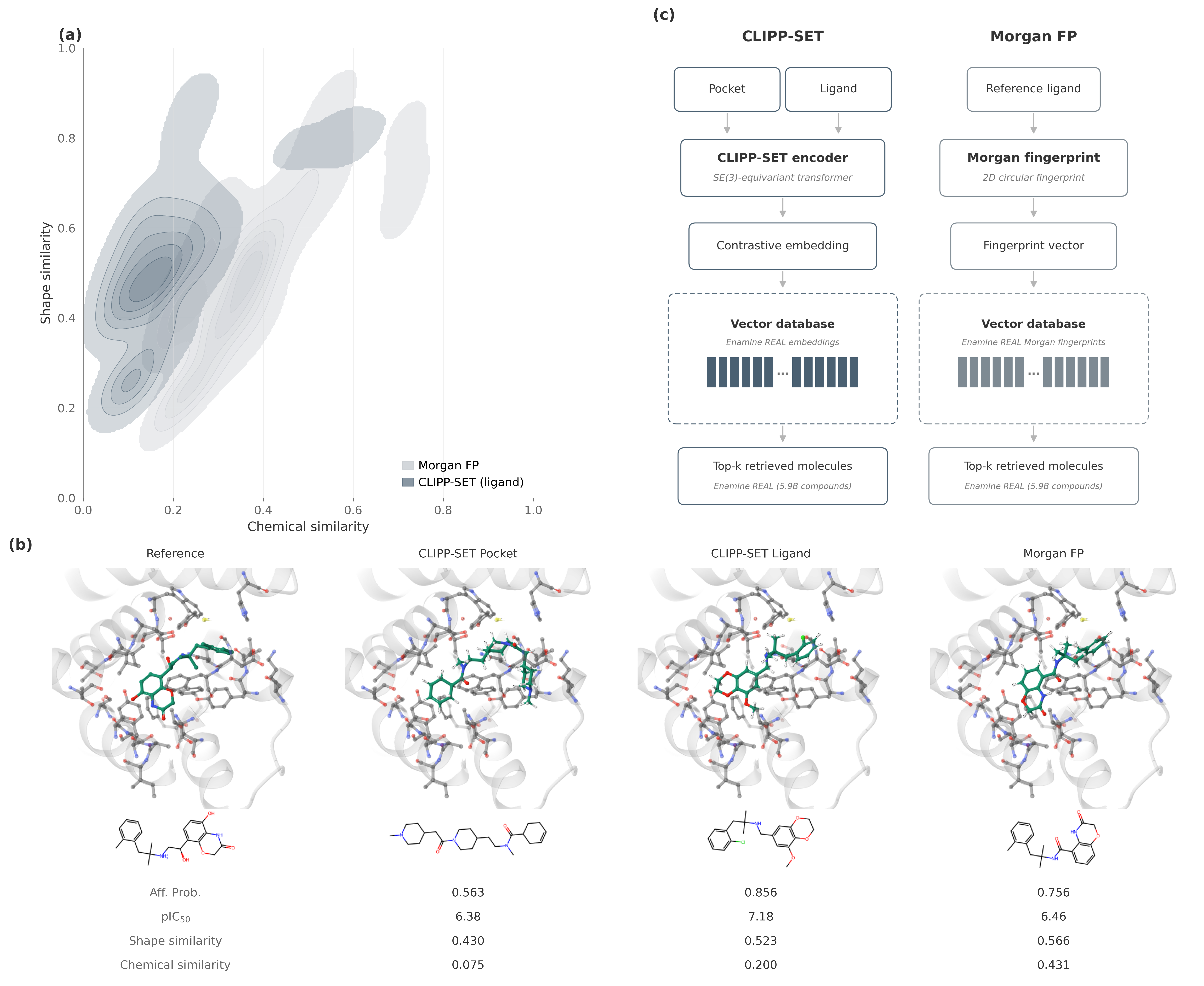}
    \caption{(a) Joint distribution of 3D shape similarity and 2D chemical similarity for molecules retrieved by CLIPP-SET ligand search and Morgan FP search across 15 LIT-PCBA targets. (b) Case study for target 3p0g (ADRB2): top-ranked molecule from each search method and reference ligand in the binding pocket, with 2D structures and predicted binding metrics. (c) Search and retrieval architecture: CLIPP-SET encodes pocket and ligand 3D structures into a shared contrastive embedding space for cosine-similarity retrieval, while Morgan FP retrieves neighbors by Tanimoto similarity over 2D circular fingerprints.}
    \label{fig:search_overview}
\end{figure}

Figure~\ref{fig:search_overview}a shows the joint distribution of 3D shape similarity and 2D chemical similarity for molecules retrieved by CLIPP-SET ligand search and Morgan FP search across all 15 targets. Morgan FP search achieves comparable shape similarity to CLIPP-SET ligand search, but this is a natural consequence of retrieving close structural analogs: molecules that are chemically similar will share similar 3D shapes. The key distinction is that the CLIPP-SET distribution is shifted toward substantially lower chemical similarity while preserving high shape similarity, indicating that the contrastive embeddings capture the pharmacophoric and geometric features relevant to binding rather than overall structural resemblance. This is precisely the regime of interest for drug discovery, where the goal is to identify chemically novel scaffolds that retain the 3D properties required for target engagement.

To illustrate our results concretely, we examined the top-ranked molecule retrieved by each method for the ADRB2 target (3p0g). Figure~\ref{fig:search_overview}b shows the predicted binding poses from Boltz-2 alongside the reference ligand and 2D structures. The CLIPP-SET ligand search retrieves a molecule with a predicted pIC$_{50}$ of 7.18 and moderate shape similarity (0.523), while bearing minimal 2D resemblance to the reference (Tanimoto = 0.200). The CLIPP-SET pocket search, which requires no reference ligand, finds a chemically unrelated molecule (Tanimoto = 0.075) with a predicted pIC$_{50}$ of 6.38. The Morgan FP search retrieves the most chemically similar molecule (Tanimoto = 0.431) with a predicted pIC$_{50}$ of 6.46. 

\subsection{Structure-Conditioned de Novo Design}
\label{sec:mclm}

In this section, we evaluate the MCLM for de novo molecular generation conditioned on the 3D structure of either a protein pocket or a known ligand. Although the MCLM was trained exclusively with ligand embeddings, pocket conditioning is enabled by the shared contrastive embedding space, where both modalities are approximately aligned. For each of the 15 LIT-PCBA targets, 100 molecules were generated using either the pocket or the ligand embedding, with the Enamine dataset token. Generated molecules were scored with Boltz-2 and 3D shape similarity as in the search experiments. Table~\ref{tab:gen_search} reports aggregated results alongside the three search methods from the previous sections.

\begin{table}[H]
    \centering
    \caption{Generative and search methods on the 15 LIT-PCBA targets (100 molecules per target). Best per column in bold.}
    \label{tab:gen_search}
    \resizebox{\textwidth}{!}{%
    \begin{tabular}{lccccc}
        \toprule
        Method & Aff.\ Prob.\ $\uparrow$ & Pred.\ pIC$_{50}$ $\uparrow$ & Shape Sim.\ $\uparrow$ & Chem.\ Sim.\ $\downarrow$ & Diversity $\uparrow$ \\
        \midrule
        \multicolumn{6}{l}{\textit{Search}} \\[2pt]
        CLIPP-SET (ligand)       & 0.55 $\pm$ 0.23 & 5.69 $\pm$ 0.66 & 0.51 $\pm$ 0.16 & 0.19 $\pm$ 0.12 & 0.74 $\pm$ 0.09 \\
        CLIPP-SET (pocket)       & 0.45 $\pm$ 0.12 & 5.71 $\pm$ 0.43 & 0.40 $\pm$ 0.07 & 0.11 $\pm$ 0.02 & 0.79 $\pm$ 0.08 \\
        Morgan FP                & 0.50 $\pm$ 0.20 & 5.45 $\pm$ 0.59 & 0.49 $\pm$ 0.16 & 0.36 $\pm$ 0.14 & 0.59 $\pm$ 0.08 \\
        \midrule
        \multicolumn{6}{l}{\textit{Generative}} \\[2pt]
        MCLM (ligand)            & \textbf{0.68 $\pm$ 0.24} & \textbf{6.31 $\pm$ 1.31} & \textbf{0.57 $\pm$ 0.19} & 0.38 $\pm$ 0.19 & 0.54 $\pm$ 0.20 \\
        MCLM (pocket)            & 0.46 $\pm$ 0.11 & 5.89 $\pm$ 0.45 & 0.37 $\pm$ 0.05 & \textbf{0.11 $\pm$ 0.02} & \textbf{0.81 $\pm$ 0.03} \\
        \bottomrule
    \end{tabular}%
    }
\end{table}

\begin{figure}[H]
    \centering
    \includegraphics[width=\linewidth]{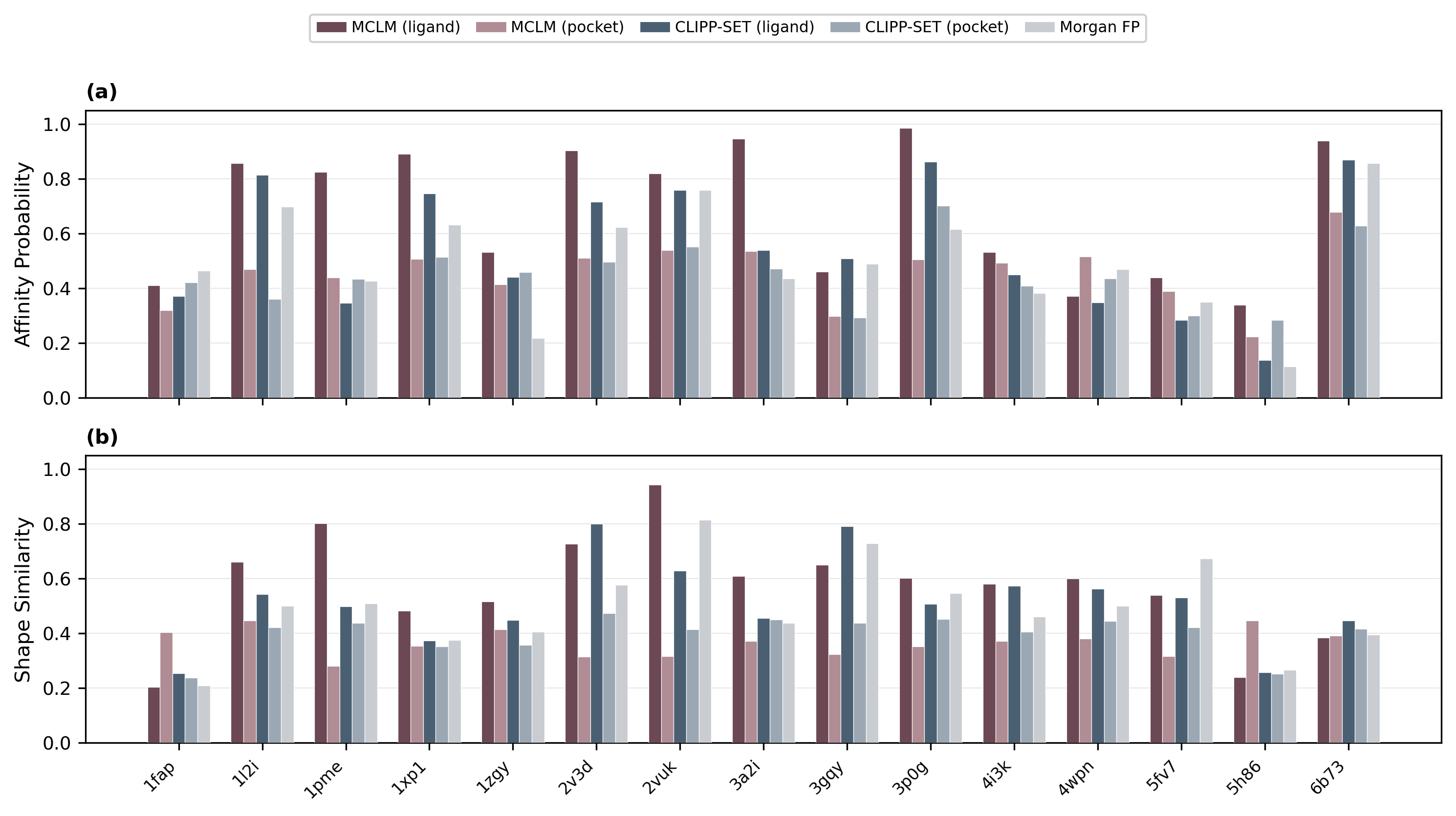}
    \caption{Per-target comparison of generative and search methods across 15 LIT-PCBA targets. (a) Predicted affinity probability. (b) 3D shape similarity to the reference ligand.}
    \label{fig:litpcba_affinity}
\end{figure}

Ligand-conditioned generation achieves the highest predicted affinity probability and pIC$_{50}$ among all methods. This advantage is consistent across individual targets (Figure~\ref{fig:litpcba_affinity}a), with ligand generation ranking first on 12 of 15 targets. Pocket-conditioned generation, which requires no reference ligand and was never explicitly trained for, reaches affinity levels comparable to the search baselines while maintaining the highest diversity of any method. The shape similarity panel (Figure~\ref{fig:litpcba_affinity}b) shows that the two ligand-based approaches, CLIPP-SET (ligand) search and MCLM (ligand) generation, exceed Morgan FP search and the pocket-based methods, with the generative model attaining the highest values overall.

The results are consistent with the contrastive embeddings capturing binding-relevant features: ligand-conditioned generation, which encodes the pharmacophoric and geometric properties of a known binder, yields the highest predicted affinity and shape similarity, while pocket-conditioned generation achieves competitive results without requiring any reference ligand, producing the most diverse candidates with minimal resemblance to existing compounds. Moreover, both generative methods outperform their respective search counterparts across most metrics, as the autoregressive model samples from an implicit chemical space theoretically orders of magnitude larger than the 5.9 billion compounds in the Enamine REAL catalog.

\subsubsection{Steering Generation Toward Targeted Chemical Spaces}
\label{sec:purchasability}

The MCLM is designed to generate molecules that resemble compounds from a target chemical space, so that candidates are more likely to be purchasable or synthesizable through established routes. The dataset token mechanism enables this control: during training, each token is associated with one of the source databases, allowing the model to learn the chemical distribution of each source. At inference time, the structural embedding and the dataset token act as complementary conditioning signals, steering generation toward the learned distribution of the desired chemical space while maintaining target specificity.

\begin{figure}[H]
    \centering
    \includegraphics[width=\linewidth]{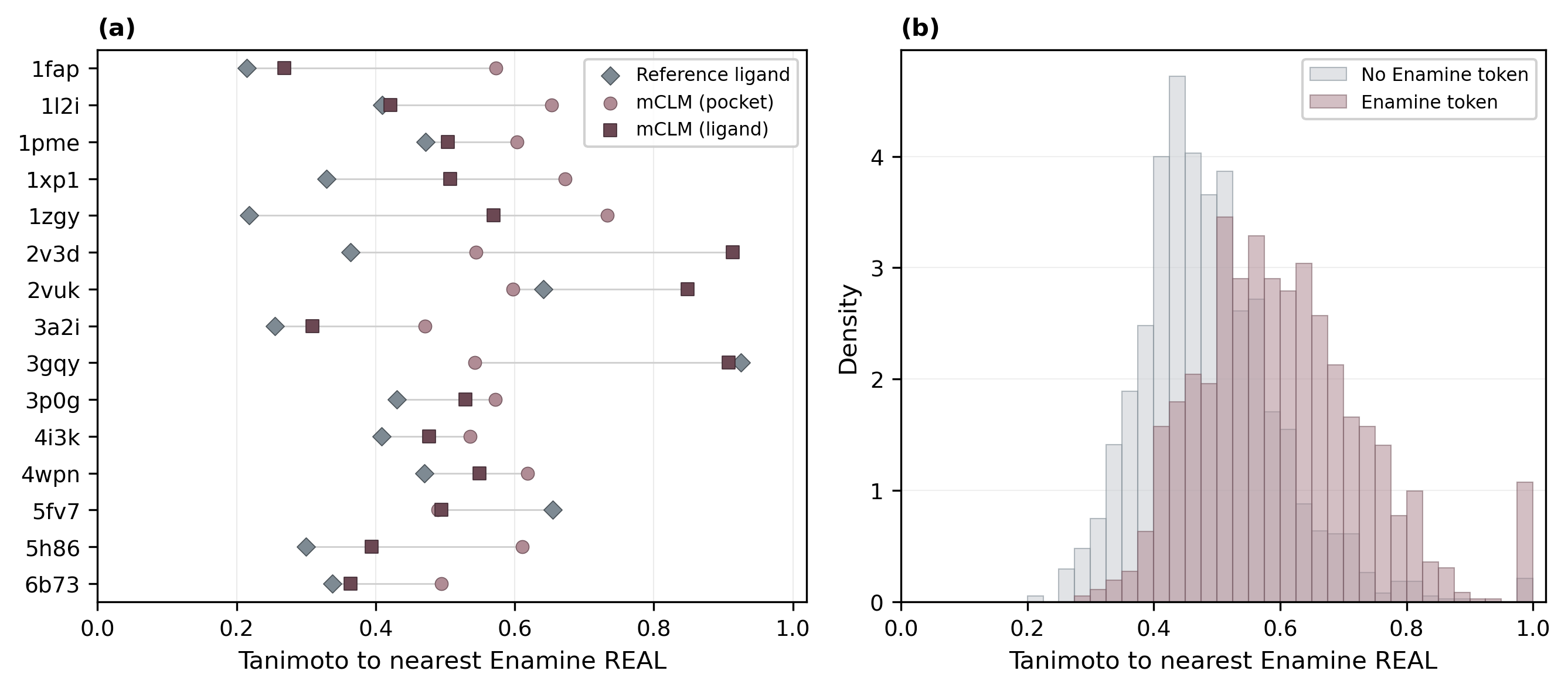}
    \caption{Chemical space steering of generated molecules. (a) Per-target Tanimoto similarity to the nearest Enamine REAL compound for the two generative conditions and the reference ligand. (b) Distribution of nearest-neighbor similarity for pocket-conditioned molecules generated with and without the Enamine token.}
    \label{fig:enamine_steering}
\end{figure}

We quantified this effect by measuring, for every generated molecule, the Tanimoto similarity to its nearest neighbor in the Enamine REAL catalog. Figure~\ref{fig:enamine_steering}a reports per-target results for both generative conditions alongside the reference ligand. In nearly all cases, generated molecules are more similar to the Enamine catalog than the reference ligand itself. Pocket-conditioned generation consistently produces molecules close to the Enamine distribution. Ligand-conditioned generation shows more variability, as the model balances two competing signals: reproducing the chemistry of the conditioning ligand and steering toward the Enamine space. When the reference ligand is already chemically close to Enamine, the model generates molecules that fall almost entirely within the catalog. When the reference ligand is more distant, the steering is harder, yet the model still redirects generation toward the target chemical space.

To isolate the contribution of the dataset token, we performed an ablation where the same pocket embeddings are decoded with and without it (Figure~\ref{fig:enamine_steering}b). Since the conformer dataset contains a substantial proportion of Enamine compounds, the model generates molecules with some similarity to the Enamine catalog even in the absence of the token, generating even some exact matches. Activating the Enamine token shifts the distribution toward higher similarity and substantially increases the fraction of exact matches, confirming that the token acts as an effective steering signal on top of the baseline distribution learned during training.

%%%%%%%%%%%%%%%%%%%%%%%%%%%%%%%%%%%%%%%%%%%%%%%%%%%%%%%%%%%%%%%%%%%%%
%% Conclusions
%%%%%%%%%%%%%%%%%%%%%%%%%%%%%%%%%%%%%%%%%%%%%%%%%%%%%%%%%%%%%%%%%%%%%
\section{Conclusions}
\label{sec:conclusions}
We presented a framework for structure-based drug design that combines a contrastive SE(3)-equivariant encoder with a multimodal Chemical Language Model. The contrastive encoder maps ligand and pocket 3D structures into a shared embedding space, enabling both zero-shot virtual screening with strong early enrichment and large-scale retrieval over billion-compound catalogs. Pocket-based retrieval identifies diverse candidates with favorable predicted binding properties using only the target structure, while ligand-based retrieval recovers molecules that match the 3D geometry of a known active with distinct chemical scaffolds.

The generative model conditions autoregressive molecular generation on these embeddings. Conditioning on ligand structures produces the strongest predicted binders across all methods evaluated, while conditioning on pocket structures alone achieves results competitive with approaches that require a known reference ligand. A learned dataset token steers the output toward a target chemical distribution, increasing the likelihood that generated molecules are purchasable or synthesizable. The framework provides a practical tool for the targeted exploration of commercial chemical spaces, uniting 3D structural specificity with the flexibility of autoregressive generation.

%%%%%%%%%%%%%%%%%%%%%%%%%%%%%%%%%%%%%%%%%%%%%%%%%%%%%%%%%%%%%%%%%%%%%
%% Acknowledgements
%%%%%%%%%%%%%%%%%%%%%%%%%%%%%%%%%%%%%%%%%%%%%%%%%%%%%%%%%%%%%%%%%%%%%
\begin{acknowledgement}
{\hfuzz=0.5pt
The project PID2020-116564GB-I00 has been funded by MCIN/AEI/10.13039/501100011033 (G.D.F.). We also acknowledge the support of the Industrial Doctorates Plan of the Secretariat of Universities and Research of the Department of Economy and Knowledge of the Generalitat of Catalonia.
\par}
\end{acknowledgement}

%%%%%%%%%%%%%%%%%%%%%%%%%%%%%%%%%%%%%%%%%%%%%%%%%%%%%%%%%%%%%%%%%%%%%
%% Supporting Information
%%%%%%%%%%%%%%%%%%%%%%%%%%%%%%%%%%%%%%%%%%%%%%%%%%%%%%%%%%%%%%%%%%%%%
\begin{suppinfo}

Supporting Information includes detailed descriptions of the Scalable Equivariant Transformer (SET) architecture, training hyperparameters for all stages of the pipeline, per-target search results on the Enamine REAL database, and per-target similarity distributions for all 15 LIT-PCBA targets.

\end{suppinfo}

%%%%%%%%%%%%%%%%%%%%%%%%%%%%%%%%%%%%%%%%%%%%%%%%%%%%%%%%%%%%%%%%%%%%%
%% Bibliography
%%%%%%%%%%%%%%%%%%%%%%%%%%%%%%%%%%%%%%%%%%%%%%%%%%%%%%%%%%%%%%%%%%%%%
\bibliography{references}

\end{document}